\documentclass[10pt,twocolumn,letterpaper]{article}

\usepackage{3dv}
\usepackage{times}
\usepackage{array}
\usepackage{epsfig}
\usepackage{graphicx}
\usepackage{amsmath,amssymb} 
\usepackage{color}
\usepackage{xcolor}
\usepackage{caption}
\usepackage{floatrow}
\usepackage[resetlabels]{multibib}
\usepackage[pagebackref=true,breaklinks=true,letterpaper=true,colorlinks,bookmarks=false]{hyperref}

\usepackage{paralist}
\usepackage{wrapfig,lipsum,booktabs}
\usepackage{makecell, pict2e}
\usepackage{pdfpages}

\newcommand{\smpl}[0]{M}
\newcommand{\posefun}[0]{T}
\newcommand{\blendfun}[0]{W}
\newcommand{\offsetfun}[0]{B}
\newcommand{\jointfun}[0]{J}

\newcommand{\pose}[0]{\boldsymbol{\theta}}
\newcommand{\shape}[0]{\boldsymbol{\beta}}

\newcommand{\blendweights}[0]{\mathbf{W}}

\newcommand{\templatetpose}[0]{\mathbf{T}_\mu}

\newcommand{\errjoints}[0]{err_{joints3D}}
\newcommand{\pckh}[0]{PCKh}

\newcommand{\errquat}[0]{err_{quat}}

\newcommand{\losslat}[0]{L_{lat}}
\newcommand{\lossthreed}[0]{L_{3D}}
\newcommand{\losstwod}[0]{L_{2D}}

\newcolumntype{x}[1]{>{\centering\arraybackslash\hspace{-9pt}}p{#1}}
\newcolumntype{y}[1]{>{\centering\arraybackslash\hspace{-6pt}}p{#1}}
\newfloatcommand{capbtabbox}{table}[][\FBwidth]

\ifthreedvfinal\fi

\newcites{apndx}{References}

\begin{document}	
\title{Neural Body Fitting: Unifying Deep Learning and Model-Based Human Pose and Shape Estimation}

\author{Mohamed Omran$^{1}$\\
\and Christoph Lassner$^{2}$\footnotemark[1]\\
\and Gerard Pons-Moll$^{1}$\\
\and Peter V. Gehler$^{2}$\footnotemark[1]\\
\and Bernt Schiele$^{1}$\\
$^{1}$Max Planck Institute for Informatics, Saarland Informatics Campus, Saarbr\"ucken, Germany\\
$^{2}$Amazon, T\"ubingen, Germany\\
{\tt\small \{mohomran, gpons, schiele\}@mpi-inf.mpg.de}, {\tt\small \{classner, pgehler\}@amazon.com}\thanks{This work was done while Christoph Lassner and Peter V. Gehler were with the MPI for Intelligent Systems and the University of T\"ubingen, and additionally while Peter was with the University of W\"urzburg.}\\
}

\maketitle

\begin{figure*}[t]
\caption{Given a single 2D image of a person, we predict a semantic body part segmentation. This part segmentation is represented as a color-coded map and used to predict the parameters of a 3D body model.}
\includegraphics[width=\textwidth]{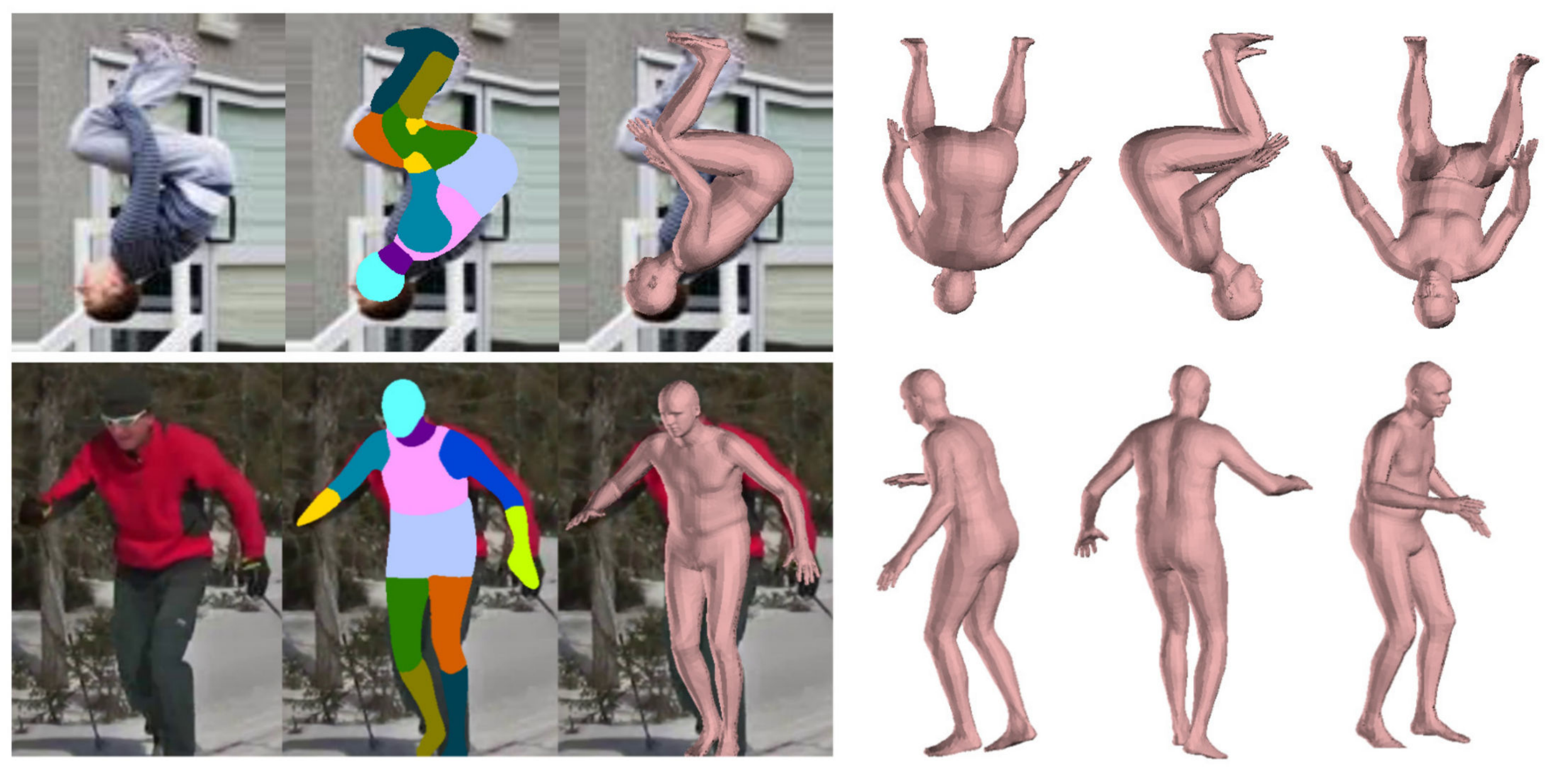}
\end{figure*}

\begin{abstract}
Direct prediction of 3D body pose and shape remains a challenge even for
highly parameterized deep learning models. Mapping from the 2D image space to 
the prediction space is difficult: perspective ambiguities make the loss
function noisy and training data is scarce. In this paper, we propose a novel 
approach (\emph{Neural Body Fitting} (NBF)).
It integrates a statistical body model within a CNN, leveraging reliable 
bottom-up semantic body part segmentation and robust top-down body model constraints. 
NBF is fully differentiable and can be trained using 2D and 3D annotations. 
In detailed experiments, we analyze how the components of our model affect performance, especially
the use of part segmentations as an explicit intermediate representation, and present a robust, efficiently trainable 
framework for 3D human pose estimation from 2D images with competitive results on standard benchmarks. 
Code will be made available at 
\url{http://github.com/mohomran/neural_body_fitting}
\end{abstract}

\vspace{-1em}

\section{Introduction}

Much research effort has been successfully directed towards predicting 
3D keypoints and stick-figure representations from images of people. 
Here, we consider the more challenging problem of estimating the parameters 
of a detailed statistical human body model from a single image. 

We tackle this problem by incorporating a model of the human body into 
a deep learning architecture, which has several advantages. First, the model 
incorporates limb orientations and shape, which are required for many 
applications such as character animation, biomechanics and virtual 
reality. Second, anthropomorphic constraints are automatically satisfied -- for example 
limb proportions and symmetry. Third, the 3D model output is one step closer to 
a faithful 3D reconstruction of people in images. 

Traditional \emph{model-based} approaches typically optimize an objective function that 
measures how well the model fits the image observations -- for example, 2D 
keypoints~\cite{bogo_smpl_eccv16,Lassner:UP:2017}. These methods do not require 
paired 3D training data (images with 3D pose), but only work well when 
initialized close to the solution. By contrast, initialization is not required 
in forward prediction models, such as CNNs that directly predict 3D keypoints.
However many  images with 3D pose annotations are required, which are difficult to 
obtain, unlike images with 2D pose annotations. 

Therefore, like us, a few recent works have proposed hybrid CNN architectures 
that are trained using model-based loss functions~\cite{tan2017indirect,tung2017self,hmrKanazawa17,pavlakos2018humanshape}.
Specifically, from an image, a CNN predicts the parameters of the SMPL 3D 
body model~\cite{smpl2015loper}, and the model is re-projected onto the 
image to evaluate the loss function in 2D space. Consequently, 2D pose 
annotations can be used to train such architectures. While these hybrid approaches 
share similarities, they all differ in essential design choices, such as the amount 
of 3D vs 2D annotations for supervision and the input representation used to lift to 3D.

To analyze the importance of such components, we introduce Neural Body Fitting
(NBF), a framework designed to provide fine-grained control over all
parts of the body fitting process. NBF is a hybrid architecture that
integrates a statistical body model within a CNN. From an RGB image
or a semantic segmentation of the image, NBF directly predicts the parameters of
the model; those parameters are passed to SMPL to produce a 3D mesh; the joints
of the 3D mesh are then projected to the image closing the loop. Hence, NBF
admits both full 3D supervision (in the model or 3D Euclidean space) and weak 2D
supervision (if images with only 2D annotations are available). NBF combines the advantages 
of direct bottom-up methods and top-down methods. It requires neither 
initialization nor large amounts of 3D training data.

One key question we address with our study is whether to use an intermediate 
representation rather than directly lifting to 3D from the raw RGB image.
Images of humans can vary due to factors such as illumination, clothing,
and background clutter. Those effects do not necessarily correlate with pose and shape, 
thus we investigate whether a simplification of the RGB image into a
semantic segmentation of body parts improves 3D inference. We also consider
the granularity of the body part segmentation as well as segmentation quality,
and find that: \begin{inparaenum}[(i)]
       \item a color-coded 12-body-part segmentation contains sufficient information for predicting shape and pose,
       \item the use of such an intermediate representation results in competitive performance and easier, more 
       data-efficient training compared to similar methods that predict pose and shape parameters from raw RGB images,
       \item segmentation quality is a strong predictor of fit quality.
     \end{inparaenum}

We also demonstrate that only a small fraction of the training data needs to be paired with 3D annotations. 
We make use of the recent UP-3D dataset~\cite{Lassner:UP:2017} that contains 8000 training images in the wild 
along with 3D pose annotations. Larger 2D datasets exist, but UP-3D allows us to perform a controlled study.

In summary, our contribution is twofold: first we introduce NBF, which unites
deep learning-based with traditional model-based methods taking advantage of
both. Second, we provide an in-depth analysis of the necessary components to
achieve good performance in hybrid architectures and provide insights for its
real-world applicability that we believe may hold for many related methods as
well.

\section{Related Work}
Human pose estimation is a well-researched field and we focus on 3D methods; for a recent extensive
survey on the field we refer to~\cite{sarafianos_posesurvey_cviu2016}.\\
\noindent {\bf Model-based Methods.} To estimate human pose and shape from images,
model-based~\cite{PonsModelBased} works use a parametric body model or template.
Early models were based on geometric
primitives~\cite{metaxas1993shape,gavrila1996,plankers2001articulated,sigal2004tracking,stoll2011fast},
while more recent 
ones are estimated from 1000s of scans of real people, and are typically parameterized by separate body pose and
shape components~\cite{anguelov2005scape,hasler2009statistical,smpl2015loper,zuffi2015stitched,pons2015dyna}, with few exceptions, \eg, \cite{joo2018total}.
Most model-based approaches fit a model to image evidence through complex non-linear optimization, requiring careful initialization to avoid poor local minima.

To reduce the complexity of the fitting procedure, the output of
2D keypoint detection has been used as additional guidance.
The progress in 2D pose
estimation using CNNs~\cite{CPM,cao2016realtime,insafutdinov17cvpr} has contributed
significantly to the task of 3D pose estimation even in challenging in-the-wild
scenarios.
For example, the 3D parameters of SMPL~\cite{smpl2015loper} can be obtained with
reasonable accuracy by fitting it to 2D
keypoints~\cite{bogo_smpl_eccv16,Lassner:UP:2017}. However, lifting to 3D from 2D
information alone is an ambiguous problem. Adversarial learning can potentially identify plausible poses~\cite{jack2017adversarially,hmrKanazawa17}.
By contrast, the seminal works of~\cite{taylor_articulated_cvpr00,sminchisescu2003kinematic} address lifting by reasoning about kinematic depth ambiguities. 
Recently, using monocular video and geometric reasoning, accurate and detailed 3D shape, including clothing is obtained~\cite{thiemo2018, thiemo2018_3DV}. 

\noindent {\bf Learning-Based Models.} Recent methods in this category typically
predict 3D keypoints or stick figures from a single image using a CNN.
High capacity models are trained on standard 3D
datasets~\cite{ionescu_human36_pami14,sigal_humaneva_ijcv10}, which are limited
in terms of appearance variation, pose, backgrounds and occlusions.
Consequently, it is not clear -- despite excellent performance on standard
benchmarks --
how methods~\cite{tekin_structured_bmvc16,li_accv14,pavlakos_volumetric_cvpr17,li_maximum_iccv2015}
generalize to in-the-wild images. To add variation, some methods resort to generating synthetic 
images ~\cite{rogez_mocap_nips16,varol17,Lassner:GP:2017} but it is complex to approximate 
fully realistic images with sufficient variance.
%
Similar to model-based methods, learning approaches have benefited from the
advent of robust 2D pose methods -- by matching 2D detections to a 3D pose
database~\cite{chen_2d_match_cvpr17,yasin_dual_source_cvpr16}, by regressing
pose from 2D joint distance matrices~\cite{moreno_distance_matrix_cvpr17}, by
exploiting pose and geometric priors for
lifting~\cite{zhou_sparseness_deepness_cvpr15,akhter_pose_conditioned_cvpr15,simo_joint_CVPR2013,Jahangiri2017,mehta2017single,zhou2018monocap,rogez2018lcr};
or simply by training a feed forward network to directly predict 3D pose from 2D
joints~\cite{martinez20173dbaseline}. Another way to exploit images with only 2D
data is by re-using the first layers of a 2D pose CNN for the task of 3D pose
estimation~\cite{tekin_fusion_arxiv16,VNect_SIGGRAPH2017,mehta_mono_3dv17}. Pavlakos et al.~\cite{pavlakos2018ordinal} take another approach by relying on weak 3D supervision in form of a relative 3D ordering of joints, similar to the previously proposed PoseBits~\cite{posebits_cvpr14}.

Closer to ours are approaches that train using separate 2D and 3D
losses~\cite{popa2017deep,zhou2017towards,sun2017compositional}. 
However, since they do not integrate a statistical
body model in the network, limbs and body proportions might be
unnatural. Most importantly, they only predict 3D stick
figures as opposed to a full mesh.

Some works regress correspondences to a body model which are then used to fit the model to depth data~\cite{pons2015metric,taylor2012vitruvian}. 
Recently, correspondences to the SMPL body surface are regressed from images directly~\cite{guler2018densepose} by leveraging dense keypoint annotations; 
however, the approach can not recover 3D human pose and shape. \cite{varol2018bodynet} fits SMPL to CNN volumetric outputs as a post-process step.
3D model fitting within a CNN have been proposed for faces~\cite{tewari2017mofa}; faces however, are not articulated like bodies which simplifies the regression problem.

A few recent works (concurrent to ours) integrate the SMPL~\cite{smpl2015loper} model within a network~\cite{tung2017self,hmrKanazawa17,pavlakos2018humanshape}. 
The approaches differ primarily in the proxy representation used to lift to 3D: 
RGB images~\cite{hmrKanazawa17}, images and 2D keypoints~\cite{tung2017self} and 2D keypoints and silhouettes~\cite{pavlakos2018humanshape}, 
and the kind of supervision (3D vs 2D) used for training. 
In contrast to previous work, we analyze the importance of such components for good performance. Kanazawa et al.~\cite{hmrKanazawa17} also integrate a learned prior on the space of poses.
We draw inspiration from model-based and learning approaches in several aspects of our model design. 
Firstly, NBF addresses an important limitation: it does not require an initialization for optimization because it incorporates a CNN based bottom-up component. 
Furthermore, at test time, NBF predictions are fast and do not require optimization. 
By integrating SMPL directly into the CNN, we do not require multiple network heads to backpropagate 2D and 3D losses. 
Lastly, we use a semantic segmentation as proxy representation, which (1) abstracts away irrelevant image information for 3D pose, (2) is a richer semantic representation than keypoints or silhouettes, and (3) allows us to analyze the importance of part granularity and placement for 3D prediction.


\section{Method}

\begin{figure*}

  \includegraphics[width=\textwidth]{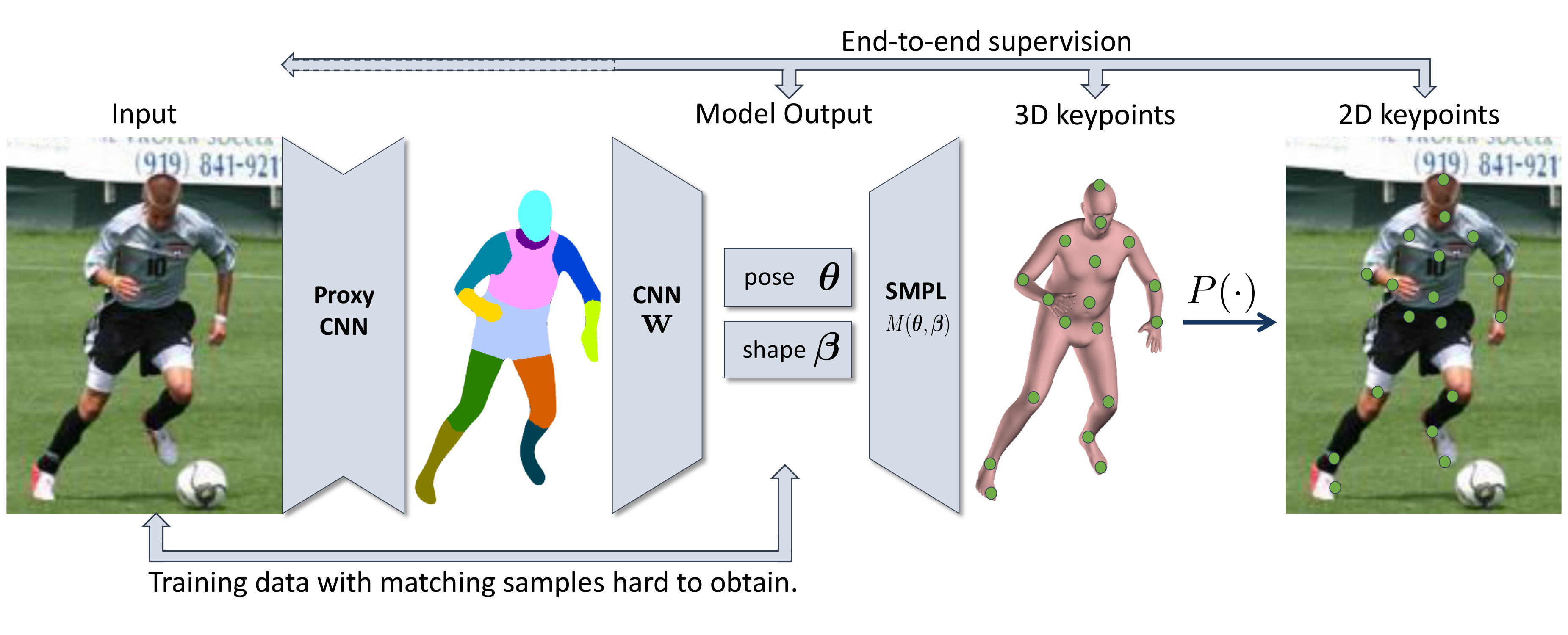}
  \vspace*{-.8cm}
  \caption{\textit{Summary of our proposed pipeline.} We process the image with a standard semantic segmentation CNN into 12 semantic parts (see Sec.~\ref{sec:impl-details}). An encoding CNN processes the semantic part probability maps to predict SMPL body model parameters (see Sec.~\ref{sec:neural-body-fitting-params}). We then use our SMPL implementation in Tensorflow to obtain a projection of the pose-defining points to 2D. With these points, a loss on 2D vertex positions can be back propagated through the entire model (see Sec.~\ref{sec:loss-functions}).}
  \label{fig:summary}

\end{figure*}

Our goal is to fit a 3D mesh of a human to a single static image (see Figure ~\ref{fig:summary}). This task involves multiple steps and we want to apply 3D but also 2D losses due to the strongly varying difficulty to obtain ground truth data. Nevertheless, we aim to build a simple processing pipeline with parts that can be optimized in isolation and avoiding multiple network heads. This reduces the number of hyperparameters and interactions, \eg, loss weights, and allows us to consecutively train the model parts. There are two main stages in the proposed architecture: in a first stage, a body part segmentation is predicted from the RGB image. The second stage takes this segmentation to predict a low-dimensional parameterization of a body mesh.\\

\vspace{-1em}

\subsection{Body Model}
\label{subsec:body_model}

For our experiments we use the SMPL body model due to its good trade-off between high anatomic flexibility and realism. SMPL parameterizes a triangulated mesh with $N=6890$ vertices with pose parameters $\theta\in\mathbb{R}^{72}$ and shape parameters $\shape \in \mathbb{R}^{10}$ -- optionally the translation parameters $\gamma\in \mathbb{R}^{3}$ can be taken into account as well.
Shape $\offsetfun_s(\shape)$ and pose dependent deformations $\offsetfun_p(\pose)$ are first applied to a base template $\templatetpose$; then the mesh is posed by rotating each body part around skeleton joints $\jointfun(\shape)$ using a skinning function $\blendfun$:
\begin{equation}
\smpl(\shape,\pose) = \blendfun(\posefun(\shape,\pose), \jointfun(\shape), \pose, \blendweights),
\label{eq:smpl}
\end{equation}
\begin{equation}
\posefun(\shape,\pose) = \templatetpose + \offsetfun_s(\shape) + \offsetfun_p(\pose),
\end{equation}
where $\smpl(\shape,\pose)$ is the SMPL function, and $\posefun(\shape,\pose)$ outputs an intermediate mesh in a T-pose after pose and shape deformations are applied. SMPL produces realistic results using relatively simple mathematical operations -- most importantly for us SMPL is fully differentiable with respect to pose $\pose$ and shape $\shape$. All these operations, including the ones to determine projected points of a posed and parameterized 3D body, are available in Tensorflow. We use them to make the 3D body a part of our deep learning model.

\subsection{Neural Body Fitting Parameterization}\label{sec:neural-body-fitting-params}

NBF predicts the parameters of the body model from a colour-coded part segmentation map $\mathbf{I} \in \mathbb{R}^{{224}\times{224}\times{3}}$ using a CNN-based predictor parameterized by weights ~$w$. The estimators for pose and shape are thus given by $\pose(w,\mathbf{I})$ and $\shape(w,\mathbf{I})$ respectively. 

We integrate the SMPL model and a simple 2D projection layer into our CNN estimator, as described in Sec.~\ref{subsec:body_model}. 
This allows us to output a 3D mesh, 3D skeleton joint locations or 2D joints, depending on the kind of supervision we want to apply for training while keeping the CNN monolithic.

Mathematically, the function $N_{3D}(w,\mathbf{I})$ that maps from semantic images to meshes is given by
\begin{eqnarray}
N_{3D}(w,\mathbf{I}) &= &M(\pose(w,\mathbf{I}),\shape(w,\mathbf{I})) \\
                     &=& \blendfun(\posefun(\shape(w,\mathbf{I}),\pose(w, \mathbf{I}),\nonumber\\
  & & \;\;\;\;\;\jointfun(\shape(w,\mathbf{I})), \pose(w,\mathbf{I}), \blendweights)),
\end{eqnarray}
which is the SMPL Equation~\eqref{eq:smpl} parameterized by network parameters $w$. 
NBF can also predict the 3D joints $N_{J}(w,\mathbf{I})=\jointfun(\shape(w,\mathbf{I})$, because they are a function of the model parameters. Furthermore, using a projection operation $\pi(\cdot)$ we can project the 3D joints onto the image plane
\begin{equation}
N_{2D}(w,\mathbf{I}) = \pi(J(w,\mathbf{I})),
\end{equation}
where $N_{2D}(w,\mathbf{I})$ is the NBF function that outputs 2D joint locations. All of these operations are differentiable and allow us to use gradient-based optimization to update model parameters with a suitable loss function.

\subsection{Loss Functions}\label{sec:loss-functions}

We experiment with the following loss functions:\\
{\em 3D latent parameter loss: } This is an L1 loss on the model parameters $\pose$ and $\shape$. Given a paired dataset $\{\mathbf{I}_i, \pose_i,\shape_i\}_i^N$, the loss is given by:
\begin{equation}
\mathcal{L}_{lat}(w) = \sum_i^{N} |\mathbf{r}(\pose(w,\mathbf{I}_i))-\mathbf{r}(\pose_i)| +  |\shape(w,\mathbf{I}_i)-\shape_i|,
\end{equation}
where $\mathbf{r}$ are the vectorized rotation matrices of the 24 parts of the body. Similar to~\cite{Lassner:UP:2017, pavlakos2018humanshape}, we observed better performance by imposing the loss on the rotation matrix representation of $\pose$ rather than on its `native' axis angle encoding as defined in SMPL. This requires us to project the predicted matrices to the manifold of rotation matrices. We perform this step using SVD to maintain differentiability.
\\
{\em 3D joint loss: }  Given a paired dataset with skeleton annotations $\{\mathbf{I}_i, \pose_i,\mathbf{J}\}_i^N$ we compute the loss in terms of 3D joint position differences as:
\begin{equation}
\mathcal{L}_{3D}(w) = \sum_i^{N} \| N_{J}(w,\mathbf{I}_i) - \mathbf{J}_i \|^2
\end{equation}
{\em 2D joint loss:} If the dataset $\{\mathbf{I}_i, \mathbf{J}_{2D}\}_i^N$ provides solely 2D joint position ground truth, we define a similar loss in terms of 2D distance and rely on error backpropagation through the projection:
\begin{equation}
\mathcal{L}_{2D}(w) = \sum_i^{N} \| N_{2D}(w,\mathbf{I}_i) - \mathbf{J}_{2D,i} \|^2
\end{equation}
{\em Joint 2D and 3D loss: }
To maximize the amounts of usable training data, ideally multiple data sources can be combined with a subset of the data $\mathcal{D}_{3D}$ providing 3D annotations and another subset $\mathcal{D}_{3D}$ providing 2D annotations. We can trivially integrate all the data with different kinds of supervision by falling back to the relevant losses and setting them to zero if not applicable.
\begin{equation}
\mathcal{L}_{2D+3D}(w,\mathcal{D}) = \mathcal{L}_{2D}(w, \mathcal{D}_{2D}) + \mathcal{L}_{3D}(w,\mathcal{D}_{3D})
\end{equation}
 In our experiments, we analyze the performance of each loss and their combinations. 
In particular, we evaluate how much gain in 3D estimation accuracy can be obtained from weak 2D annotations which are much cheaper to obtain than accurate 3D annotations. 

\section{Results}
\subsection{Evaluation Settings}

We used the following three datasets for evaluation: UP-3D,
\cite{Lassner:UP:2017},  HumanEva-I \cite{sigal_humaneva_ijcv10} 
and Human3.6M\cite{ionescu_human36_pami14}.
We perform a detailed analysis of our approach on UP-3D and Human3.6M, 
and compare against state-of-the-art methods on HumanEVA-I and Human3.6M.

UP-3D, is a challenging, in-the-wild dataset that draws from existing pose
datasets: LSP \cite{johnson_lsp_bmvc10}, LSP-extended \cite{johnson_lsp_bmvc10},
MPII HumanPose \cite{andriluka_mpii2d_cvpr14}, and FashionPose \cite{dantone14fashionpose}. It
augments images from these datasets with rich 3D annotations in the form of SMPL
model parameters that fully capture shape and pose, allowing us to derive 2D
and 3D joint as well as fine-grained segmentation annotations. The dataset consists of training 
(5703 images), validation (1423 images) and test (1389 images) sets. 
For our analysis, we use the training set and provide results on the validation set.

The HumanEVA-I dataset is recorded in a controlled environment with marker-based ground truth synchronized with video.
The dataset includes 3 subjects and 2 motion sequences per subject. Human3.6M also consists of similarly recorded data
but covers more subjects, with 15 action sequences per subject repeated over two trials. For our analysis
on this dataset, we reserve subjects S1, S5, S6 and S7 for training, holding out subject S8 for validation. 
We compare to the state of art on the test sequences S9 and S11.

\subsection{Implementation Details}\label{sec:impl-details}

\textbf{Data preparation:} To train our model, we require images paired with 3D body model fits 
(i.e. SMPL parameters) as well as pixelwise part labels. The UP-3D dataset provides such annotations, 
while Human3.6M does not. However, by applying MoSH \cite{Loper:SIGASIA:2014} 
to the 3D MoCap marker data provided by the latter we obtain the corresponding SMPL parameters,
which in turn allows us to generate part labels by rendering an appropriately annotated SMPL mesh \cite{Lassner:UP:2017}. 

\textbf{Scale ambiguity:} The SMPL shape parameters encode among other factors a person's size. 
Additionally, both distance to the camera and focal length determine how large a person appears in an image.
To eliminate this ambiguity during training, we constrain scale information to the shape parameters by making the following 
assumptions: The camera is always at the SMPL coordinate origin, the optical axis always points in the same direction, and a person
is always at a fixed distance from the camera. We render the ground truth SMPL fits and scale the training images to fit the renderings (using the corresponding 2D joints). 
This guarantees that the the only factor affecting person size in the image are the SMPL shape parameters. At test-time, 
we estimate person height and location in the image using 2D DeeperCut keypoints~\cite{insafutdinov_deepercut_eccv16},
and center the person within a 512x512 crop such that they have a height of 440px, which roughly corresponds to the setting seen during training.

\textbf{Architecture:} We use a two-stage approach: The first stage receives the 512x512 input crop and produces a part
segmentation. We use a RefineNet~\cite{lin17refinenet} model (based on ResNet-101 \cite{he_resnet_cvpr2016}). 
This part segmentation is color-coded, resized to 224x224 and fed as an RGB image to the second stage, 
itself composed of two parts: a regression network (ResNet-50) that outputs the 226 SMPL parameters (shape and pose), 
and a non-trainable set of layers that implement the SMPL model and an image projection. Such layers can produce a 3D mesh, 3D joints or 2D joints
given the predicted pose and shape. Training both stages requires a total of 18 (12+6) hours on a single Volta V100 GPU.
More details are provided in the supplementary material as well as in the code (to be released). 

\subsection{Analysis}

\paragraph{\textbf{Which Input Encoding?}}

We investigate here what input representation is effective for pose and shape prediction. 
Full RGB images certainly contain more information than for example silhouettes, part segmentations or 2D joints. 
However, some information may not be relevant for 3D inference, such as appearance, illumination or clothing, 
which might make the network overfit to nuisances factors 

To this end, we train a network on different image representations and compare their performance on the UP-3D and
Human3.6M validation sets. We compare RGB images, color-coded part segmentations of 
varying granularities, and color-coded joint heatmaps (see supplementary material for examples). 
We generate both using the ground truth SMPL annotations to establish an upper bound
on performance, and later consider the case where we do not have access to such
information at test time.

\begin{table}
\caption{\textit{Input Type vs. 3D error in millimeters}}\label{tab:inptype}
\begin{tabular}{l||x{1.5cm}x{1.5cm}}\toprule  
  \emph{type of input}          & \emph{UP}      & \emph{H36M}        \tabularnewline   
  \hline
  \hline
    RGB                         & 98.5           & 48.9                 \\\midrule
    Segmentation (1 part)       & 95.5           & 43.0                 \\
    Segmentation (3 parts)      & 36.5           & 37.5                 \\
    Segmentation (6 parts)      & 29.4           & 36.2                 \\
    Segmentation (12 parts)     & 27.8           & 33.5                 \\
    Segmentation (24 parts)     & 28.8           & 31.8                 \\\midrule
    Joints (14)                 & 28.8           & 33.4                 \\
    Joints (24)                 & 27.7           & 33.4                 \\
\end{tabular}
\end{table}

The results are reported in Table~\ref{tab:inptype}. We observe that explicit part representations (part segmentations or joint heatmaps) are more useful for
3D shape/pose estimation compared to RGB images and plain silhouettes. The difference is especially pronounced on the UP-3D dataset, which contains more visual 
variety than the images of Human3.6M, with an error drop from 98.5 mm to 27.8 mm when using a 12 part segmentation. 
This demonstrates that a 2D segmentation of the person into sufficient parts carries a lot of information about 3D pose/shape,
while also providing full spatial coverage of the person (compared to joint heatmaps).
Is it then worth learning separate mappings first from image to part segmentation, and then 
from part segmentation to 3D shape/pose? To answer this question we first need to examine how 3D accuracy is affected by the quality of real predicted part segmentations.

\begin{figure}
  \includegraphics[width=1.0\textwidth]{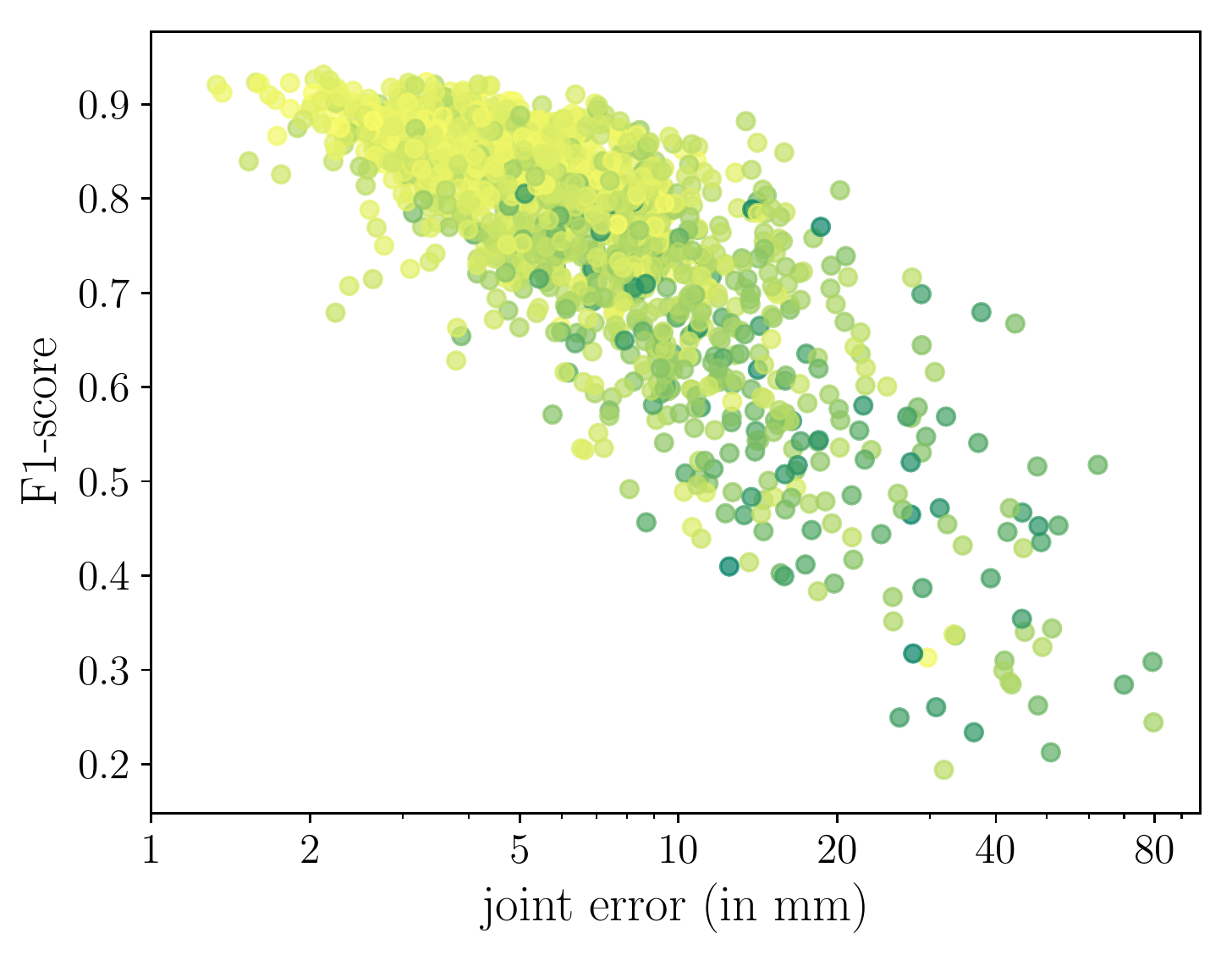}
  
  \vspace{-1.5em}
  \caption{\textit{Segmentation quality (F1-score) vs. fit quality (3D joint error).} The darkness indicates the difficulty of the pose, i.e. the distance from the upright pose with arms by the sides. 
  \label{fitsegm}}%
\end{figure}

\textbf{Which Input Quality?} To determine the effect of segmentation quality on the results, we train three different \emph{part segmentation networks}.
Besides RefineNet, we also train two variants of DeepLab~\cite{chen2016deeplab}, based on VGG-16~\cite{simonyan2014very} and ResNet-101~\cite{he_resnet_cvpr2016}. 
These networks result in IoU scores of $67.1$, $57.0$, and $53.2$ 
respectively on the UP validation set. Given these results, we then train four \emph{3D prediction networks} - one for each of the part segmentation networks, and 
an additional one using the ground truth segmentations. We report 3D accuracy on the validation set of UP3D for each of the four 3D networks, diagonal numbers of Table~\ref{tab:Effect-of-training-set}.
As one would expect, the better the segmentation, the better the 3D prediction accuracy. 
As can also be seen in Table~\ref{tab:Effect-of-training-set}, better segmenters at test time always lead to improved 3D accuracy, even when the 3D prediction networks are trained with poorer segmenters.  
This is perhaps surprising, and it indicates that mimicking the statistics of a particular segmentation method at training time plays only a minor role (for example a network trained with GT segmentations and tested using RefineNet segmentations performs comparably to a network that is trained using RefineNet segmentations (83.3mm vs 82mm)). 
To further analyze the correlation between segmentation quality and 3D accuracy, in Figure~\ref{fitsegm} we plot the relationship 
between F-1 score and 3D reconstruction error. Each dot represents one image, and the color its respective difficulty -- we use the distance to mean pose as a proxy measure for difficulty.
The plot clearly shows that the higher the F-1 score, the lower the 3D joint error.

\begin{table}
  {\setlength{\extrarowheight}{1pt}
  \begin{tabular}{c|cccc} \toprule
  \diaghead{\hskip5.2em}{Val}{Train} & VGG & ResNet & RefineNet & GT \tabularnewline
  \hline
  \hline
  VGG         &  $107.2$         &  $119.9$         &  $135.5$         &  $140.7$            \tabularnewline
  ResNet      &  $97.1 $         &  $96.3 $         &  $112.2$         &  $115.6$            \tabularnewline
  RefineNet      &  $89.6 $         &  $89.9 $         &  $82.0 $         &  $83.3 $            \tabularnewline
  GT             &  $62.3 $         &  $60.5 $         &  $35.7 $         &  $27.8 $            \tabularnewline
  \end{tabular}}
  \caption{\label{tab:Effect-of-training-set}\textit{Effect of segmentation quality on the quality of the 3D fit prediction modules ($\errjoints$)}}%
  \vspace{-0.5em}
\end{table}

\textbf{Which Types of Supervision?} We now examine different combinations of loss terms.
The losses we consider are $\losslat$ (on the latent parameters), $\lossthreed$ (on 3D joint/vertex locations), 
$\losstwod$ (on the projected joint/vertex locations).
We compare performance using three different error measures: 
     \begin{inparaenum}[(i)]
       \item $\errjoints$, the Euclidean distance between ground truth and predicted SMPL joints (in mm).
       \item $\pckh$ \cite{andriluka_mpii2d_cvpr14}, the percentage of correct keypoints with the error threshold being $50\%$ of head size, which we measure on a per-example basis.
       \item $\errquat$, quaternion distance error of the predicted joint rotations (in radians).
     \end{inparaenum}

Given sufficient data - the full 3D-annotated UTP training set with mirrored examples (11406) - only applying a loss on the 
model parameters yields reasonable results, and in this setting, additional loss terms don't provide benefits. \\
When only training with $\lossthreed$, we obtain similar results in terms of $\errjoints$, however, interestingly $\errquat$
is significantly higher. This indicates that predictions produce accurate 3D joints positions in space, but the limb orientations are incorrect. 
This further demonstrates that methods trained to produce only 3D keypoints do not capture orientation, which is needed for many applications.

We also observe that only training with the 2D reprojection loss (perhaps unsurprisingly) 
results in poor performance in terms of 3D error, showing that some amount of 3D annotations are necessary to overcome 
the ambiguity inherent to 2D keypoints as a source of supervision for 3D. 

Due to the SMPL layers, we can supervise learning with any number of joints/mesh vertices. We thus experimented with the 91 landmarks used by~\cite{Lassner:UP:2017} for their fitting method but find that the 24 SMPL joints are sufficient in this setting.
\begin{table}
  \caption{\textit{Loss ablation study.} Results in 2D and 3D error metrics (\textit{joints3D}: Euclidean 3D distance, \textit{mesh}: average vertex to vertex distance, \textit{quat}: average body part rotation error in radians).}\label{wrap-tab:1}
\begin{center}
\small
\begin{tabular*}{1.0\textwidth}{l||x{1.9cm}x{1.05cm}x{1.2cm}}\toprule  
\emph{Loss}                 & $\mathbf{\errjoints}$   & $\mathbf{\pckh}$     & $\mathbf{\errquat} $  \\\midrule
$L_{lat}$                   & 83.7                    & 93.1                 & 0.278                 \\
$L_{lat} + L_{3D}$          & 82.3                    & 93.4                 & 0.280                 \\  
$L_{lat} + L_{2D}$          & 83.1                    & 93.5                 & 0.278                 \\
$L_{lat} + L_{3D} + L_{2D}$ & 82.0                    & 93.5                 & 0.279                 \\
$L_{3D}$                    & 83.7                    & 93.5                 & 1.962                 \\
$L_{2D}$                    & 198.0                   & 94.0                 & 1.971                 \\
\end{tabular*}
\end{center}
\end{table}

\vspace{-0.2em}
\textbf{How Much 3D Supervision Do We Need?} The use of these additional loss terms also allows us to leverage data for which 
no 3D annotations are available. With the following set of experiments,
we attempt to answer two questions: (i) Given a small amount of 3D-annotated 
data, does extra 2D-annotated data help?, (ii) What amount of 3D data is necessary? 
To this end we train multiple networks, each time progressively disabling the 3D latent 
loss and replacing it with the 2D loss for more training examples. The results are depicted in Table \ref{tab:more2d}. 
We find that performance barely degrades as long as we have a small amount of 3D annotations. 
In contrast, using small amounts of 3D data and no extra data with 2D annotations yields poor performance. 
This is an important finding since obtaining 3D annotations is difficult compared to simple 2D keypoint annotations.
\begin{center}
\begin{table}
  \caption{\textit{Effect of 3D labeled data.} We show the 3D as well as the estimated body part rotation error for varying ratios of data with 3D labels. For all of the data, we assume that 2D pose labels are available. Both errors saturate at 20\% of 3D labeled training examples.}\label{wrap-tab:1}
  \begin{center}
\small
\label{tab:more2d}
\begin{tabular*}{1.0\textwidth}{l||y{0.4cm}y{0.4cm}y{0.4cm}y{0.4cm}y{0.4cm}y{0.4cm}y{0.4cm}y{0.4cm}}\toprule  
\diaghead{\hskip5.2em}{Error}{{Ann.perc.}}& $100$& $50$ & $20$ & $10$ & $5$  & $2$ & $1$ & $0$ \\\midrule
$\mathbf{\errjoints}$                                & 83.1 & 82.8 & 82.8   & 83.6   & 84.5   & 88.1  & 93.9  & 198 \\
$\mathbf{\errquat}  $                                & 0.28 & 0.28 & 0.27   & 0.28   & 0.29   & 0.30  & 0.33  & 1.97  \\  
\end{tabular*}
\end{center}
\end{table}
\end{center}
\begin{figure*}
   \caption{\textit{Qualitative results by error quartile in terms of $\errjoints$.} 
   The rows show representative examples from different error quartiles, top to bottom: 0-25\%, 25-50\%, 50-75\%, 75-100\%} 
   \includegraphics[width=\textwidth]{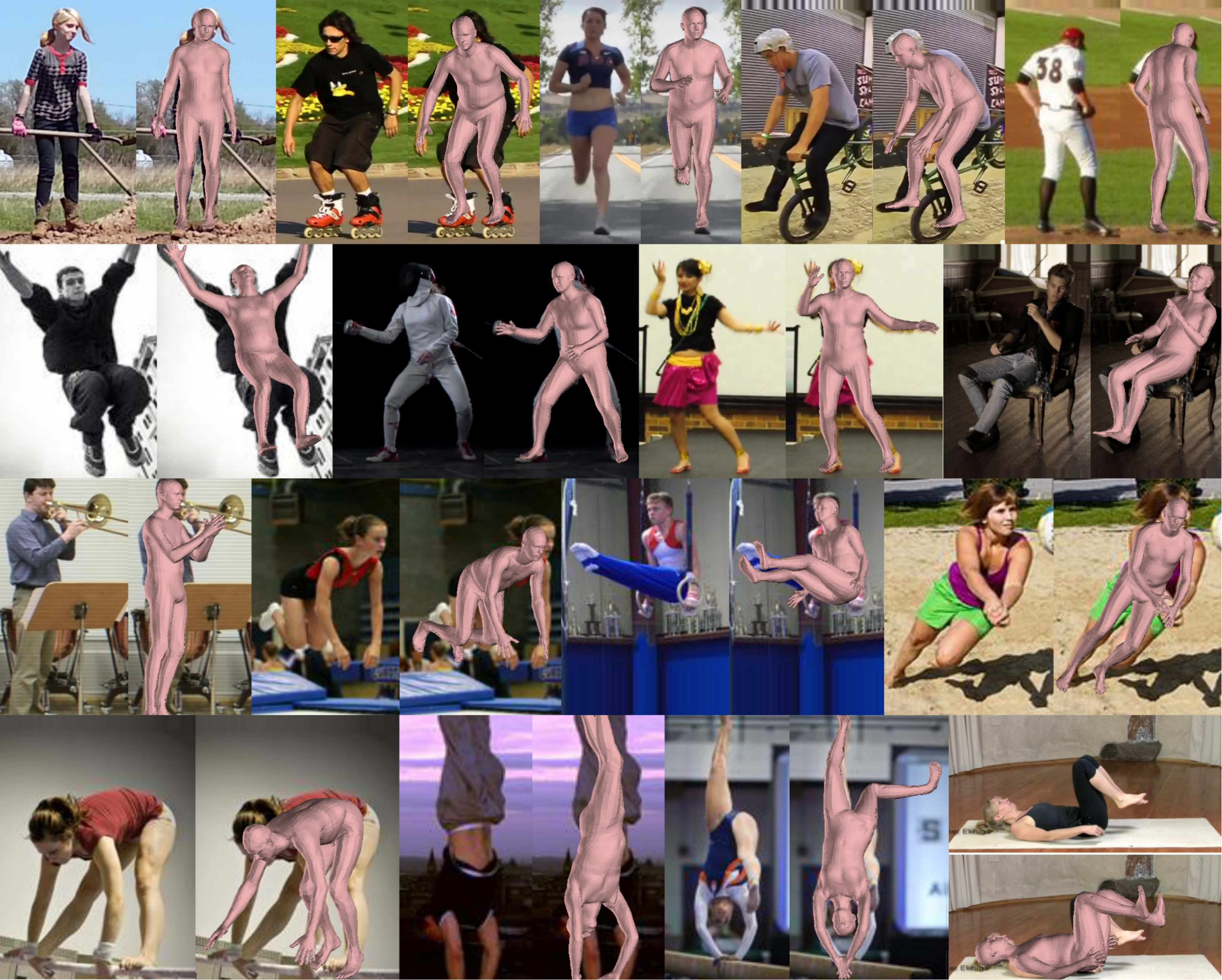}
    \vspace{-1.5em}
   \label{fig:quali}
\end{figure*}
\vspace{-0.2em}
\textbf{Qualitative Results} A selection of qualitative results from the UP-3D dataset can be found in Figure ~\ref{fig:quali}. 
We show examples from the four different error quartiles. Fits from the first three quartiles still reproduce 
the body pose somewhat faithfully, and only in the last row and percentile, problems become clearly visible. 
We show more failure modes in the supplementary material. 

\vspace{-0.3em}

\subsection{Comparison to State-of-the-Art}
Here we compare to the state of the art on HumanEva-I (Table~\ref{table:heva}) and Human3.6M (Table~\ref{table:H36}).
We perform a per-frame rigid alignment of the 3D estimates to the ground truth using Procrustes Analysis
and report results in terms of reconstruction error, i.e. the mean per joint position error after alignment 
(given in $mm$). The model we use here is trained on Human3.6M data.

\begin{table}[t]
\centering
\begin{tabular}{lrr}
\hline
Method & Mean & Median \\
\hline
Ramakrishna et al. \cite{ramakrishna2012reconstructing}   & 168.4 & 145.9\\
Zhou et al.        \cite{zhou_convexrelaxation_cvpr2015}  & 110.0 &  98.9\\
SMPLify            \cite{bogo_smpl_eccv16}                &  79.9 &  61.9\\
Random Forests     \cite{Lassner:UP:2017}                 &  93.5 &  77.6\\
SMPLify (Dense)    \cite{Lassner:UP:2017}                 &  74.5 &  59.6\\
Ours                                                      &  64.0 &  49.4\\
\hline
\end{tabular}
\caption{{\bf HumanEva-I results.} 3D joint errors in mm. }
\label{table:heva}
\end{table}

\begin{table}
\begin{tabular}{lrrrrrrrrr}
\hline
 Method & \multicolumn{1}{c}{Mean}      & \multicolumn{1}{l}{Median} \\ \hline
 Akhter \& Black \cite{akhter_pose_conditioned_cvpr15}        & 181.1                         & 158.1               \\
 Ramakrishna et al. \cite{ramakrishna2012reconstructing}      & 157.3                         & 136.8               \\
 Zhou et al. \cite{zhou_convexrelaxation_cvpr2015}            & 106.7                         &  90.0               \\
 SMPLify         \cite{bogo_smpl_eccv16}                      &  82.3                         &  69.3               \\
 SMPLify (dense) \cite{Lassner:UP:2017}                       &  80.7                         &  70.0               \\ \hline
 SelfSup \cite{tung2017self}                                  &  98.4                         &    -                \\
 Pavlakos et al. \cite{pavlakos2018humanshape}                &  75.9                         &    -                \\
 HMR (H36M-trained)\cite{hmrKanazawa17}                       &  77.6                         &  72.1               \\
 HMR               \cite{hmrKanazawa17}                       &  \bf{56.8}                    &    -                \\
 Ours                                                         &  59.9                         &  52.3               \\ \hline
 \end{tabular}
 \caption{{\bf Human 3.6M.} {3D joint errors in mm}.
 }
 \label{table:H36}
 \end{table}

We compare favourably to similar methods, but these are not strictly 
comparable since they train on different datasets. Pavlakos et al. \cite{pavlakos2018humanshape} 
do not use any data from Human3.6M, whereas HMR \cite{hmrKanazawa17} does, along with several other datasets.
We retrained the latter with the original code only using Human3.6M data for a more direct comparison to ours (HMR (H36M-trained) in Table~\ref{table:H36}). 
Given Table~\ref{tab:inptype}, we hypothesize that their approach 
requires more training data for good performance because it uses RGB images as input.

\section{Conclusion}

In this paper, we make several principled steps towards a full integration of
parametric 3D human pose models into deep CNN architectures. We analyze (1) how
the 3D model can be integrated into a deep neural network, (2) how loss
functions can be combined and (3) how a training can be set up that works 
efficiently with scarce 3D data.

In contrast to existing methods we use a region-based 2D representation,
namely a 12-body-part segmentation, as an intermediate step prior to the 
mapping to 3D shape and pose. This segmentation provides full spatial coverage 
of a person as opposed to the commonly used sparse set of keypoints, while 
also retaining enough information about the arrangement of parts to allow
for effective lifting to 3D.

We used a stack of CNN layers on top of a segmentation model to predict
an encoding in the space of 3D model parameters, followed by a Tensorflow
implementation of the 3D model and a projection to the image plane. This full
integration allows us to finely tune the loss functions and enables end-to-end
training. We found a loss that combines 2D as well as 3D information to work best.
The flexible implementation allowed us to experiment with the 3D losses only for
parts of the data, moving towards a weakly supervised training scenario that
avoids expensive 3D labeled data. With 3D information for only 20\% of our
training data, we could reach similar performance as with full 3D annotations.

We believe that this encouraging result is an important finding for the
design of future datasets and the development of 3D prediction methods
that do not require expensive 3D annotations for training. Future work
will involve extending this to more challenging settings involving
multiple, possibly occluded, people. \vspace{0.1em}

\textbf{Acknowledgements} We would like to thank Dingfan Chen for help with training the HMR \cite{hmrKanazawa17} model.

\clearpage
\bibliographystyle{ieee}
\bibliography{egbib}

\begin{thebibliography}{1}\itemsep=-1pt

\bibitem{BelagiannisRCN15}
V.~Belagiannis, C.~Rupprecht, G.~Carneiro, and N.~Navab.
\newblock Robust optimization for deep regression.
\newblock 2015.

\bibitem{GemanMcclure1987}
S.~Geman and D.~McClure.
\newblock Statistical methods for tomographic image reconstruction.
\newblock {\em Bulletin of the International Statistical Institute},
  52(4):5--21, 1987.

\bibitem{kingma15adam}
D.~P. Kingma and J.~Ba.
\newblock Adam: {A} method for stochastic optimization.
\newblock 2014.

\bibitem{lin17refinenet2}
G.~Lin and I.~R. Anton~Milan, Chunhua~Shen.
\newblock Refinenet: Multi-path refinement networks for high-resolution
  semantic segmentation.
\newblock In {\em Computer Vision and Pattern Recognition (CVPR)}, 2017.

\end{thebibliography}


\begin{thebibliography}{10}\itemsep=-1pt

\bibitem{akhter_pose_conditioned_cvpr15}
I.~Akhter and M.~J. Black.
\newblock Pose-conditioned joint angle limits for 3d human pose reconstruction.
\newblock In {\em IEEE Conference on Computer Vision and Pattern Recognition
  (CVPR)}, pages 1446--1455, 2015.

\bibitem{thiemo2018_3DV}
T.~Alldieck, M.~Magnor, W.~Xu, C.~Theobalt, and G.~Pons-Moll.
\newblock Detailed human avatars from monocular video.
\newblock In {\em 3D Vision (3DV), 2018 Sixth International Conference on},
  2018.

\bibitem{thiemo2018}
T.~Alldieck, M.~Magnor, W.~Xu, C.~Theobalt, and G.~Pons-Moll.
\newblock Video based reconstruction of 3d people models.
\newblock In {\em {IEEE} Conference on Computer Vision and Pattern
  Recognition}, June 2018.

\bibitem{andriluka_mpii2d_cvpr14}
M.~Andriluka, L.~Pishchulin, P.~Gehler, and B.~Schiele.
\newblock {2D Human Pose Estimation: New Benchmark and State of the Art
  Analysis}.
\newblock In {\em IEEE Conference on Computer Vision and Pattern Recognition
  (CVPR)}, June 2014.

\bibitem{anguelov2005scape}
D.~Anguelov, P.~Srinivasan, D.~Koller, S.~Thrun, J.~Rodgers, and J.~Davis.
\newblock Scape: shape completion and animation of people.
\newblock In {\em ACM Transactions on Graphics (TOG)}, volume~24, pages
  408--416. ACM, 2005.

\bibitem{bogo_smpl_eccv16}
F.~Bogo, A.~Kanazawa, C.~Lassner, P.~Gehler, J.~Romero, and M.~J. Black.
\newblock Keep it {SMPL}: Automatic estimation of {3D} human pose and shape
  from a single image.
\newblock In {\em European Conference on Computer Vision (ECCV)}, 2016.

\bibitem{cao2016realtime}
Z.~Cao, T.~Simon, S.-E. Wei, and Y.~Sheikh.
\newblock Realtime multi-person 2d pose estimation using part affinity fields.
\newblock In {\em Proc. of CVPR}, 2016.

\bibitem{chen_2d_match_cvpr17}
C.-H. Chen and D.~Ramanan.
\newblock 3d human pose estimation = 2d pose estimation + matching.
\newblock In {\em CVPR 2017-IEEE Conference on Computer Vision \& Pattern
  Recognition}, 2017.

\bibitem{chen2016deeplab}
L.-C. Chen, G.~Papandreou, I.~Kokkinos, K.~Murphy, and A.~L. Yuille.
\newblock Deeplab: Semantic image segmentation with deep convolutional nets,
  atrous convolution, and fully connected crfs.
\newblock {\em arXiv preprint arXiv:1606.00915}, 2016.

\bibitem{dantone14fashionpose}
M.~Dantone, J.~Gall, C.~Leistner, and L.~V. Gool.
\newblock Body parts dependent joint regressors for human pose estimation in
  still images.
\newblock {\em IEEE Transactions on Pattern Analysis and Machine Intelligence},
  36(11):2131--2143, Nov 2014.

\bibitem{gavrila1996}
D.~M. Gavrila and L.~S. Davis.
\newblock 3-d model-based tracking of humans in action: a multi-view approach.
\newblock In {\em Proc. CVPR}, pages 73--80. IEEE, 1996.

\bibitem{guler2018densepose}
R.~A. G{\"u}ler, N.~Neverova, and I.~Kokkinos.
\newblock Densepose: Dense human pose estimation in the wild.
\newblock {\em arXiv preprint arXiv:1802.00434}, 2018.

\bibitem{hasler2009statistical}
N.~Hasler, C.~Stoll, M.~Sunkel, B.~Rosenhahn, and H.-P. Seidel.
\newblock A statistical model of human pose and body shape.
\newblock In {\em Computer Graphics Forum}, volume~28, pages 337--346, 2009.

\bibitem{he_resnet_cvpr2016}
K.~He, X.~Zhang, S.~Ren, and J.~Sun.
\newblock Deep residual learning for image recognition.
\newblock In {\em IEEE Conference on Computer Vision and Pattern Recognition
  (CVPR)}, 2016.

\bibitem{insafutdinov17cvpr}
E.~Insafutdinov, M.~Andriluka, L.~Pishchulin, S.~Tang, E.~Levinkov, B.~Andres,
  and B.~Schiele.
\newblock Arttrack: {A}rticulated multi-person tracking in the wild.
\newblock In {\em 30th IEEE Conference on Computer Vision and Pattern
  Recognition (CVPR 2017)}, Honolulu, HI, USA, 2017. IEEE.

\bibitem{insafutdinov_deepercut_eccv16}
E.~Insafutdinov, L.~Pishchulin, B.~Andres, M.~Andriluka, and B.~Schiele.
\newblock Deepercut: A deeper, stronger, and faster multi-person pose
  estimation model.
\newblock In {\em European Conference on Computer Vision (ECCV)}, 2016.

\bibitem{ionescu_human36_pami14}
C.~Ionescu, D.~Papava, V.~Olaru, and C.~Sminchisescu.
\newblock {Human3.6m: Large scale datasets and predictive methods for 3d human
  sensing in natural environments}.
\newblock {\em IEEE Transactions on Pattern Analysis and Machine Intelligence
  (PAMI)}, 36(7):1325--1339, 2014.

\bibitem{jack2017adversarially}
D.~Jack, F.~Maire, A.~Eriksson, and S.~Shirazi.
\newblock Adversarially parameterized optimization for 3d human pose
  estimation.
\newblock 2017.

\bibitem{Jahangiri2017}
E.~Jahangiri and A.~L. Yuille.
\newblock Generating multiple diverse hypotheses for human 3d pose consistent
  with 2d joint detections.
\newblock In {\em IEEE International Conference on Computer Vision (ICCV)
  Workshops (PeopleCap)}, 2017.

\bibitem{johnson_lsp_bmvc10}
S.~Johnson and M.~Everingham.
\newblock Clustered pose and nonlinear appearance models for human pose
  estimation.
\newblock In {\em British Machine Vision Conference (BMVC)}, 2010.
\newblock doi:10.5244/C.24.12.

\bibitem{joo2018total}
H.~Joo, T.~Simon, and Y.~Sheikh.
\newblock Total capture: A 3d deformation model for tracking faces, hands, and
  bodies.
\newblock {\em arXiv preprint arXiv:1801.01615}, 2018.

\bibitem{hmrKanazawa17}
A.~Kanazawa, M.~J. Black, D.~W. Jacobs, and J.~Malik.
\newblock End-to-end recovery of human shape and pose.
\newblock In {\em Proceedings of the IEEE Conference on Computer Vision and
  Pattern Recognition}, 2018.

\bibitem{Lassner:GP:2017}
C.~Lassner, G.~Pons-Moll, and P.~V. Gehler.
\newblock A generative model of people in clothing.
\newblock In {\em International Conference on Computer Vision (ICCV)}, 2017.

\bibitem{Lassner:UP:2017}
C.~Lassner, J.~Romero, M.~Kiefel, F.~Bogo, M.~J. Black, and P.~V. Gehler.
\newblock Unite the people: Closing the loop between 3d and 2d human
  representations.
\newblock In {\em IEEE Conf. on Computer Vision and Pattern Recognition
  (CVPR)}, 2017.

\bibitem{li_accv14}
S.~Li and A.~B. Chan.
\newblock 3d human pose estimation from monocular images with deep
  convolutional neural network.
\newblock In {\em Asian Conference on Computer Vision (ACCV)}, pages 332--347,
  2014.

\bibitem{li_maximum_iccv2015}
S.~Li, W.~Zhang, and A.~B. Chan.
\newblock Maximum-margin structured learning with deep networks for 3d human
  pose estimation.
\newblock In {\em IEEE International Conference on Computer Vision (ICCV)},
  pages 2848--2856, 2015.

\bibitem{lin17refinenet}
G.~Lin and I.~R. Anton~Milan, Chunhua~Shen.
\newblock Refinenet: Multi-path refinement networks for high-resolution
  semantic segmentation.
\newblock In {\em Computer Vision and Pattern Recognition (CVPR)}, 2017.

\bibitem{smpl2015loper}
M.~Loper, N.~Mahmood, J.~Romero, G.~Pons-Moll, and M.~J. Black.
\newblock {SMPL}: A skinned multi-person linear model.
\newblock {\em ACM Trans. Graphics (Proc. SIGGRAPH Asia)}, 34(6):248:1--248:16,
  Oct. 2015.

\bibitem{Loper:SIGASIA:2014}
M.~M. Loper, N.~Mahmood, and M.~J. Black.
\newblock {MoSh}: Motion and shape capture from sparse markers.
\newblock {\em ACM Transactions on Graphics, (Proc. SIGGRAPH Asia)},
  33(6):220:1--220:13, Nov. 2014.

\bibitem{martinez20173dbaseline}
J.~Martinez, R.~Hossain, J.~Romero, and J.~J. Little.
\newblock A simple yet effective baseline for 3d human pose estimation.
\newblock In {\em IEEE International Conference on Computer Vision (ICCV)},
  2017.

\bibitem{mehta_mono_3dv17}
D.~Mehta, H.~Rhodin, D.~Casas, P.~Fua, O.~Sotnychenko, W.~Xu, and C.~Theobalt.
\newblock Monocular 3d human pose estimation in the wild using improved cnn
  supervision.
\newblock In {\em 3D Vision (3DV), 2017 Fifth International Conference on}.
  IEEE, 2017.

\bibitem{mehta2017single}
D.~Mehta, O.~Sotnychenko, F.~Mueller, W.~Xu, S.~Sridhar, G.~Pons-Moll, and
  C.~Theobalt.
\newblock Single-shot multi-person 3d pose estimation from monocular rgb.
\newblock In {\em 3D Vision (3DV), 2018 Sixth International Conference on},
  2018.

\bibitem{VNect_SIGGRAPH2017}
D.~Mehta, S.~Sridhar, O.~Sotnychenko, H.~Rhodin, M.~Shafiei, H.-P. Seidel,
  W.~Xu, D.~Casas, and C.~Theobalt.
\newblock Vnect: Real-time 3d human pose estimation with a single rgb camera.
\newblock {\em ACM Transactions on Graphics}, 36(4), July 2017.

\bibitem{metaxas1993shape}
D.~Metaxas and D.~Terzopoulos.
\newblock Shape and nonrigid motion estimation through physics-based synthesis.
\newblock {\em IEEE Trans. PAMI}, 15(6):580--591, 1993.

\bibitem{moreno_distance_matrix_cvpr17}
F.~Moreno-Noguer.
\newblock 3d human pose estimation from a single image via distance matrix
  regression.
\newblock In {\em CVPR 2017-IEEE Conference on Computer Vision \& Pattern
  Recognition}, 2017.

\bibitem{pavlakos2018ordinal}
G.~Pavlakos, X.~Zhou, and K.~Daniilidis.
\newblock Ordinal depth supervision for 3d human pose estimation.
\newblock In {\em Proceedings of the IEEE Conference on Computer Vision and
  Pattern Recognition}, 2018.

\bibitem{pavlakos_volumetric_cvpr17}
G.~Pavlakos, X.~Zhou, K.~G. Derpanis, and K.~Daniilidis.
\newblock Coarse-to-fine volumetric prediction for single-image 3{D} human
  pose.
\newblock In {\em CVPR 2017-IEEE Conference on Computer Vision \& Pattern
  Recognition}, 2017.

\bibitem{pavlakos2018humanshape}
G.~Pavlakos, L.~Zhu, X.~Zhou, and K.~Daniilidis.
\newblock Learning to estimate 3{D} human pose and shape from a single color
  image.
\newblock In {\em Proceedings of the IEEE Conference on Computer Vision and
  Pattern Recognition}, 2018.

\bibitem{plankers2001articulated}
R.~Plankers and P.~Fua.
\newblock Articulated soft objects for video-based body modeling.
\newblock In {\em International Conference on Computer Vision, Vancouver,
  Canada}, number CVLAB-CONF-2001-005, pages 394--401, 2001.

\bibitem{posebits_cvpr14}
G.~Pons-Moll, D.~J. Fleet, and B.~Rosenhahn.
\newblock Posebits for monocular human pose estimation.
\newblock In {\em IEEE Conference on Computer Vision and Pattern Recognition
  (CVPR)}, pages 2337--2344, 2014.

\bibitem{pons2015dyna}
G.~Pons-Moll, J.~Romero, N.~Mahmood, and M.~J. Black.
\newblock Dyna: a model of dynamic human shape in motion.
\newblock {\em ACM Transactions on Graphics (TOG)}, 34:120, 2015.

\bibitem{PonsModelBased}
G.~Pons-Moll and B.~Rosenhahn.
\newblock {\em Model-Based Pose Estimation}, chapter~9, pages 139--170.
\newblock Springer, 2011.

\bibitem{pons2015metric}
G.~Pons-Moll, J.~Taylor, J.~Shotton, A.~Hertzmann, and A.~Fitzgibbon.
\newblock Metric regression forests for correspondence estimation.
\newblock {\em International Journal of Computer Vision}, 113(3):163--175,
  2015.

\bibitem{popa2017deep}
A.-I. Popa, M.~Zanfir, and C.~Sminchisescu.
\newblock Deep multitask architecture for integrated 2d and 3d human sensing.
\newblock {\em IEEE Conference on Computer Vision and Pattern Recognition
  (CVPR)}, 2017.

\bibitem{ramakrishna2012reconstructing}
V.~Ramakrishna, T.~Kanade, and Y.~Sheikh.
\newblock Reconstructing 3d human pose from 2d image landmarks.
\newblock In {\em European Conference on Computer Vision}, pages 573--586.
  Springer, 2012.

\bibitem{rogez_mocap_nips16}
G.~Rogez and C.~Schmid.
\newblock Mocap-guided data augmentation for 3d pose estimation in the wild.
\newblock In {\em Advances in Neural Information Processing Systems}, pages
  3108--3116, 2016.

\bibitem{rogez2018lcr}
G.~Rogez, P.~Weinzaepfel, and C.~Schmid.
\newblock Lcr-net++: Multi-person 2d and 3d pose detection in natural images.
\newblock {\em arXiv preprint arXiv:1803.00455}, 2018.

\bibitem{sarafianos_posesurvey_cviu2016}
N.~Sarafianos, B.~Boteanu, B.~Ionescu, and I.~A. Kakadiaris.
\newblock 3d human pose estimation: A review of the literature and analysis of
  covariates.
\newblock {\em Computer Vision and Image Understanding}, 152:1--20, 2016.

\bibitem{sigal_humaneva_ijcv10}
L.~Sigal, A.~O. Balan, and M.~J. Black.
\newblock Humaneva: Synchronized video and motion capture dataset and baseline
  algorithm for evaluation of articulated human motion.
\newblock {\em International Journal of Computer Vision (IJCV)}, 87(1-2):4--27,
  2010.

\bibitem{sigal2004tracking}
L.~Sigal, S.~Bhatia, S.~Roth, M.~J. Black, and M.~Isard.
\newblock Tracking loose-limbed people.
\newblock In {\em Computer Vision and Pattern Recognition, 2004. CVPR 2004.
  Proceedings of the 2004 IEEE Computer Society Conference on}, volume~1, pages
  I--421. IEEE, 2004.

\bibitem{simo_joint_CVPR2013}
E.~Simo-Serra, A.~Quattoni, C.~Torras, and F.~Moreno-Noguer.
\newblock A joint model for 2d and 3d pose estimation from a single image.
\newblock In {\em Conference on Computer Vision and Pattern Recognition
  (CVPR)}, pages 3634--3641, 2013.

\bibitem{simonyan2014very}
K.~Simonyan and A.~Zisserman.
\newblock Very deep convolutional networks for large-scale image recognition.
\newblock {\em arXiv preprint arXiv:1409.1556}, 2014.

\bibitem{sminchisescu2003kinematic}
C.~Sminchisescu and B.~Triggs.
\newblock Kinematic jump processes for monocular 3d human tracking.
\newblock In {\em Computer Vision and Pattern Recognition, 2003. Proceedings.
  2003 IEEE Computer Society Conference on}, volume~1, pages I--I. IEEE, 2003.

\bibitem{stoll2011fast}
C.~Stoll, N.~Hasler, J.~Gall, H.-P. Seidel, and C.~Theobalt.
\newblock Fast articulated motion tracking using a sums of gaussians body
  model.
\newblock In {\em Proc. ICCV}, pages 951--958. IEEE, 2011.

\bibitem{sun2017compositional}
X.~Sun, J.~Shang, S.~Liang, and Y.~Wei.
\newblock Compositional human pose regression.
\newblock {\em arXiv preprint arXiv:1704.00159}, 2017.

\bibitem{tan2017indirect}
J.~K.~V. Tan, I.~Budvytis, and R.~Cipolla.
\newblock Indirect deep structured learning for 3d human body shape and pose
  prediction.
\newblock In {\em BMVC}, volume~3, page~6, 2017.

\bibitem{taylor_articulated_cvpr00}
C.~J. Taylor.
\newblock Reconstruction of articulated objects from point correspondences in a
  single uncalibrated image.
\newblock In {\em IEEE Conference on Computer Vision and Pattern Recognition
  (CVPR)}, volume~1, pages 677--684, 2000.

\bibitem{taylor2012vitruvian}
J.~Taylor, J.~Shotton, T.~Sharp, and A.~Fitzgibbon.
\newblock The vitruvian manifold: Inferring dense correspondences for one-shot
  human pose estimation.
\newblock In {\em Computer Vision and Pattern Recognition (CVPR), 2012 IEEE
  Conference on}, pages 103--110. IEEE, 2012.

\bibitem{tekin_structured_bmvc16}
B.~Tekin, I.~Katircioglu, M.~Salzmann, V.~Lepetit, and P.~Fua.
\newblock {Structured Prediction of 3D Human Pose with Deep Neural Networks}.
\newblock In {\em British Machine Vision Conference (BMVC)}, 2016.

\bibitem{tekin_fusion_arxiv16}
B.~{Tekin}, P.~{M{\'a}rquez-Neila}, M.~{Salzmann}, and P.~{Fua}.
\newblock Learning to fuse 2d and 3d image cues for monocular body pose
  estimation.
\newblock In {\em IEEE International Conference on Computer Vision (ICCV)},
  2017.

\bibitem{tewari2017mofa}
A.~Tewari, M.~Zollh{\"o}fer, H.~Kim, P.~Garrido, F.~Bernard, P.~Perez, and
  C.~Theobalt.
\newblock Mofa: Model-based deep convolutional face autoencoder for
  unsupervised monocular reconstruction.
\newblock In {\em The IEEE International Conference on Computer Vision (ICCV)},
  volume~2, 2017.

\bibitem{tung2017self}
H.-Y. Tung, H.-W. Tung, E.~Yumer, and K.~Fragkiadaki.
\newblock Self-supervised learning of motion capture.
\newblock In {\em Advances in Neural Information Processing Systems}, pages
  5242--5252, 2017.

\bibitem{varol2018bodynet}
G.~Varol, D.~Ceylan, B.~Russell, J.~Yang, E.~Yumer, I.~Laptev, and C.~Schmid.
\newblock Bodynet: Volumetric inference of 3d human body shapes.
\newblock {\em arXiv preprint arXiv:1804.04875}, 2018.

\bibitem{varol17}
G.~Varol, J.~Romero, X.~Martin, N.~Mahmood, M.~J. Black, I.~Laptev, and
  C.~Schmid.
\newblock {Learning from Synthetic Humans}.
\newblock In {\em CVPR}, 2017.

\bibitem{CPM}
S.-E. Wei, V.~Ramakrishna, T.~Kanade, and Y.~Sheikh.
\newblock Convolutional pose machines.
\newblock In {\em 2015 IEEE Conference on Computer Vision and Pattern
  Recognition (CVPR)}, 2016.

\bibitem{yasin_dual_source_cvpr16}
H.~Yasin, U.~Iqbal, B.~Kr{\"u}ger, A.~Weber, and J.~Gall.
\newblock {A Dual-Source Approach for 3D Pose Estimation from a Single Image}.
\newblock In {\em Conference on Computer Vision and Pattern Recognition
  (CVPR)}, 2016.

\bibitem{zhou2017towards}
X.~Zhou, Q.~Huang, X.~Sun, X.~Xue, and Y.~Wei.
\newblock Towards 3d human pose estimation in the wild: A weakly-supervised
  approach.
\newblock In {\em Proceedings of the IEEE Conference on Computer Vision and
  Pattern Recognition}, pages 398--407, 2017.

\bibitem{zhou_convexrelaxation_cvpr2015}
X.~Zhou, S.~Leonardos, X.~Hu, and K.~Daniilidis.
\newblock {3D shape estimation from 2D landmarks: A convex relaxation
  approach}.
\newblock In {\em IEEE Conference on Computer Vision and Pattern Recognition
  (CVPR)}, pages 4447--4455, 2015.

\bibitem{zhou_sparseness_deepness_cvpr15}
X.~Zhou, M.~Zhu, S.~Leonardos, K.~Derpanis, and K.~Daniilidis.
\newblock {Sparseness Meets Deepness: 3D Human Pose Estimation from Monocular
  Video}.
\newblock In {\em IEEE Conference on Computer Vision and Pattern Recognition
  (CVPR)}, 2015.

\bibitem{zhou2018monocap}
X.~Zhou, M.~Zhu, G.~Pavlakos, S.~Leonardos, K.~G. Derpanis, and K.~Daniilidis.
\newblock Monocap: Monocular human motion capture using a cnn coupled with a
  geometric prior.
\newblock {\em IEEE Transactions on Pattern Analysis and Machine Intelligence},
  2018.

\bibitem{zuffi2015stitched}
S.~Zuffi and M.~J. Black.
\newblock The stitched puppet: A graphical model of 3d human shape and pose.
\newblock In {\em 2015 IEEE Conference on Computer Vision and Pattern
  Recognition (CVPR)}, pages 3537--3546. IEEE, 2015.

\end{thebibliography}

\appendix

\section{Further Qualitative Results}

\begin{figure*}[t]
\caption{Worst examples from the validation set in terms of 3D error given imperfect segmentations.}
\label{fig:corr1}
\includegraphics[width=\textwidth]{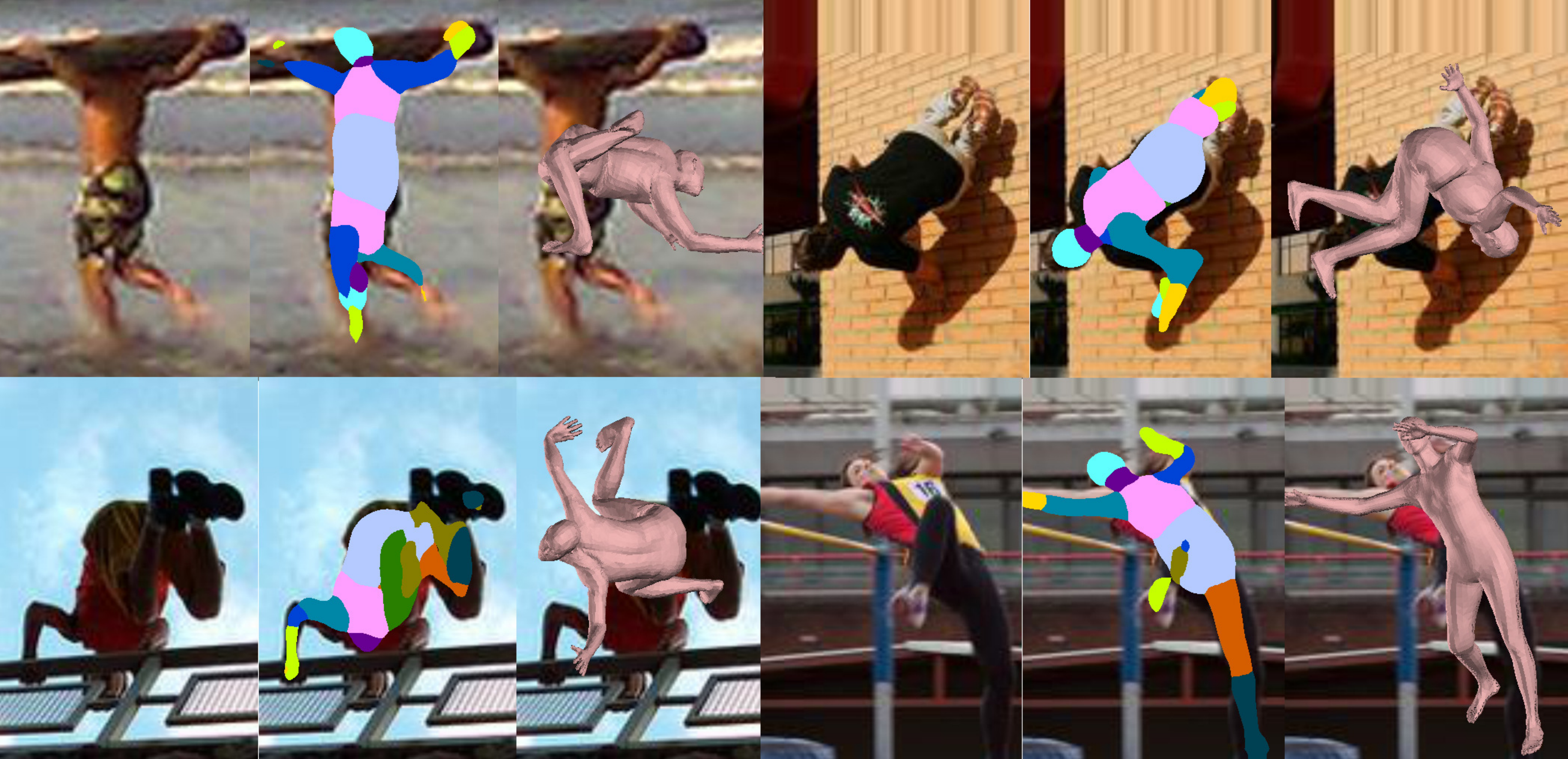}
\end{figure*}

\begin{figure*}[t]
\caption{Worst examples from the validation set in terms of 3D error given perfect segmentations.}
\label{fig:corr2}
\includegraphics[width=\textwidth]{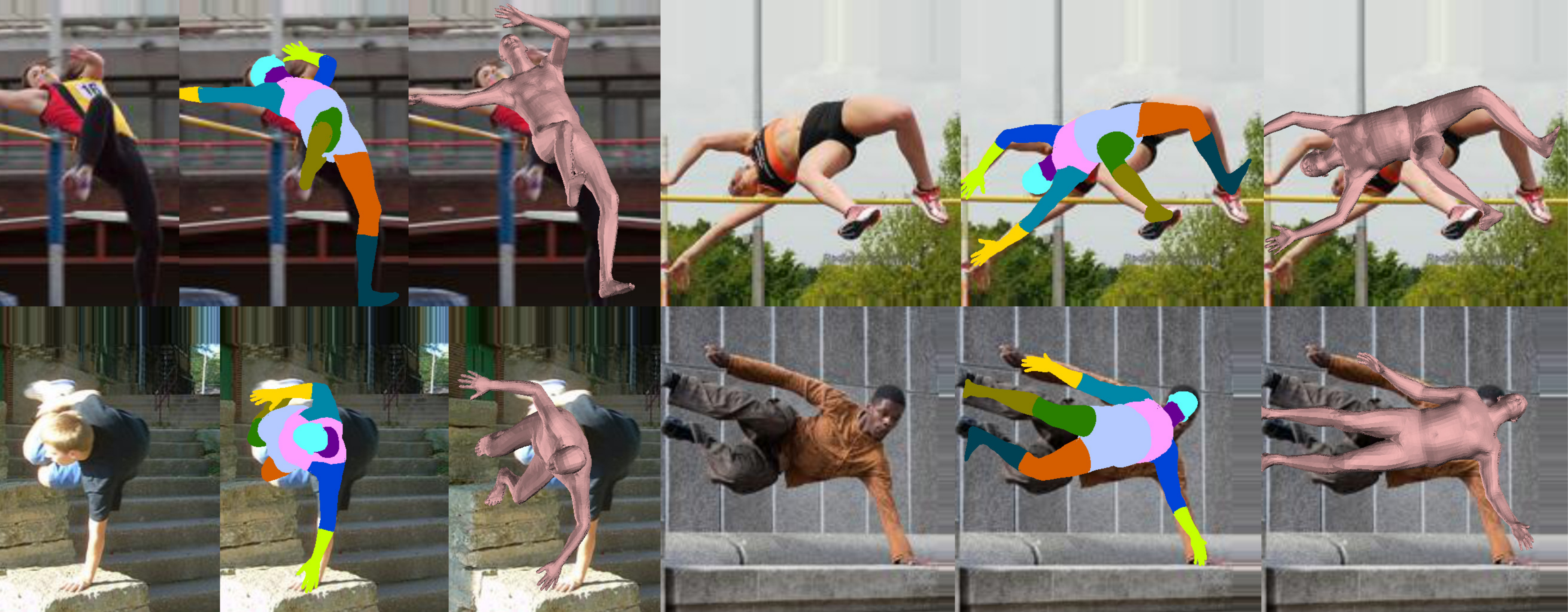}
\end{figure*}

One of our findings is the high correlation between input segmentation
quality and output fit quality. We provide some additional qualitative examples that
illustrate this correlation. In Fig.~\ref{fig:corr1}, we present the four worst examples from the
validation set in terms of 3D joint reconstruction error when we use our trained
part segmentation network; in Fig.~\ref{fig:corr2}, we present the worst examples when the
network is trained to predict body model parameters given the ground truth
segmentations. This does not correct all estimated 3D bodies, but the
remaining errors are noticeably less severe.

\section{Training Details}

\begin{figure*}[t]
\caption{Example training images annotations illustrating different types and granularities.}
\label{fig:trex}
\includegraphics[width=\textwidth]{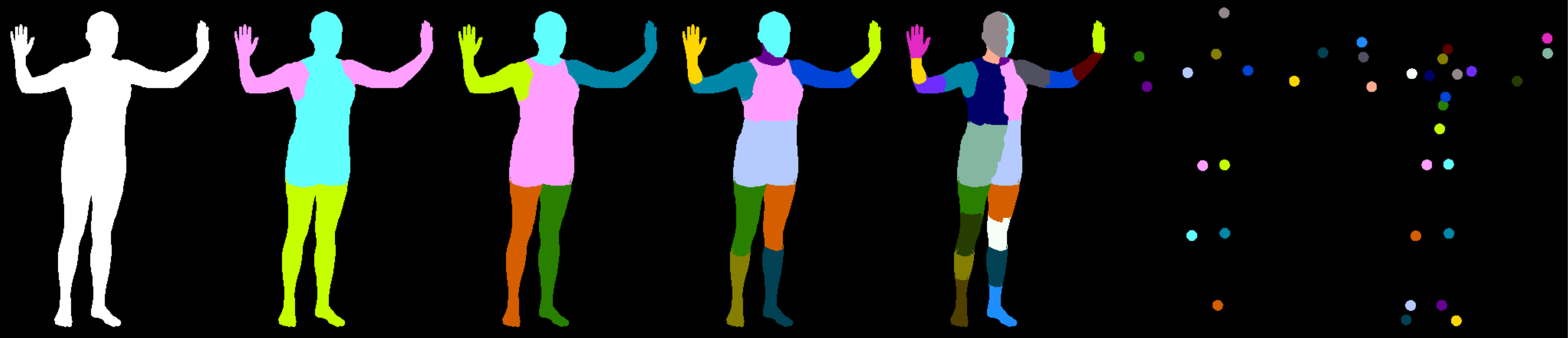}
\end{figure*}

We present examples of paired training examples and ground truth in Fig~\ref{fig:trex}.

\vspace{-0.5em}
\paragraph{Segmentation Network} We train our own TensorFlow implementation of a RefineNet~\citeapndx{lin17refinenet2} network
(based on ResNet-101) to predict the part segmentations.
The images are cropped to 512x512 pixels, and we train for 20 epochs with a batch size 
of 5 using the Adam~\citeapndx{kingma15adam} optimizer. Learning rate and weight decay are set 
to $0.00002$ and $0.0001$ respectively, with a polynomial learning rate decay. 
Data augmentation improved performance a lot, in particular horizontal 
reflection (which requires re-mapping the labels for left and right limbs), 
scale augmentation (0.9 - 1.1 of the original size) as well as rotations (up to 45 degrees). 
For training the segmentation network on UP-3D we used the 5703 training images. For Human3.6M we subsampled the videos, 
only using every 10th frame from each video, which results in about 32000 frames. 
Depending on the amount of data, training the segmentation networks takes about 6-12 hours on a Volta V100 machine.
\vspace{-0.5em}
\paragraph{Fitting Network} For the fitting network we repurpose a ResNet-50 network pretrained on ImageNet to regress
the SMPL model parameters. We replace the final pooling layer with a single fully-connected
layer that outputs the 10 shape and 216 pose parameters. We train this network for 75 epochs with a 
batch size of 5 using the Adam optimizer. The learning rate is set to 0.00004 with polynomial decay 
and we use a weight decay setting of 0.0001. We found that an L1 loss on the SMPL parameters was
a little better than an L2 loss. We also experimented with robust losses (e.g. Geman-McClure~\citeapndx{GemanMcclure1987} 
and Tukey's biweight loss~\citeapndx{BelagiannisRCN15}) but did not observe benefits. Training this network takes 
about 1.5 hours for the UP-3D dataset and six hours for Human3.6M.
\vspace{-0.5em}
\paragraph{Data Augmentation} At test-time we cannot guarantee that the person will be perfectly centered in the input crop,
which can lead to degraded performance. We found it thus critical to train both the segmentation network and the fitting network
with strong data augmentation, especially by introducing random jitter and scaling. For the fitting network, 
such augmentation has to take place prior to training since it affects the SMPL parameters. We also mirror the data, but this
requires careful mirroring of both the part labels as well as the SMPL parameters. This involves remapping the parts, 
as well as inverting the part rotations.


\bibliographystyleapndx{ieee}
\bibliographyapndx{egbib}


\end{document}